\documentclass[11pt,onecolumn]{article}
\usepackage[utf8]{inputenc}
\usepackage{amsmath,amsfonts,amssymb,graphicx,authblk,placeins,cite}
\usepackage{comment}
\usepackage{color,soul}
\usepackage{hyphenat}
\usepackage[top=50pt,bottom=50pt,left=70pt,right=70pt]{geometry}
\usepackage{sectsty}
\usepackage[font=sf]{caption}

\begin{document}
\allsectionsfont{\sffamily}
\pagenumbering{gobble}

%=========================================================================
\title{\sffamily \Huge Neural Network based Successor Representations of Space and Language
 \vspace{1 cm}}

%=========================================================================

\author[1,2]{\sffamily Paul Stoewer}
\author[1,2]{\sffamily Christian Schlieker}
\author[1,3]{\sffamily Achim Schilling}
\author[3,4]{\sffamily Claus Metzner}
\author[2]{\sffamily Andreas Maier}
\author[1,2,3,5]{\sffamily Patrick Krauss}

%=========================================================================

\affil[1]{\small{Cognitive Computational Neuroscience Group, University Erlangen-Nuremberg, Germany}}

\affil[2]{\small{Pattern Recognition Lab, University Erlangen-Nuremberg, Germany}}

\affil[3]{\small{Neuroscience Lab, University Hospital Erlangen, Germany}}

\affil[4]{\small{Biophysics Lab, University Erlangen-Nuremberg, Germany}}

\affil[5]{\small{Linguistics Lab, University Erlangen-Nuremberg, Germany}}

\maketitle

%=========================================================================

%=========================================================================
\begin{abstract}{\sffamily \noindent
How does the mind organize thoughts? The hippocampal-entorhinal complex is thought to support domain-general representation and processing of structural knowledge of arbitrary state, feature and concept spaces. In particular, it enables the formation of cognitive maps, and navigation on these maps, thereby broadly contributing to cognition. It has been proposed that the concept of multi-scale successor representations provides an explanation of the underlying computations performed by place and grid cells. Here, we present a neural network based approach to learn such representations, and its application to different scenarios: a spatial exploration task based on supervised learning, a spatial navigation task based on reinforcement learning, and a non-spatial task where linguistic constructions have to be inferred by observing sample sentences. In all scenarios, the neural network correctly learns and approximates the underlying structure by building successor representations. Furthermore, the resulting neural firing patterns are strikingly similar to experimentally observed place and grid cell firing patterns. We conclude that cognitive maps and neural network-based successor representations of structured knowledge provide a promising way to overcome some of the short comings of deep learning towards artificial general intelligence.
}
\end{abstract}

%=========================================================================
\newpage
\section*{Introduction}

Cognitive maps are mental representations that serve an organism to acquire, code, store, recall, and decode information about the relative locations and features of objects \cite{tolman1948cognitive}. Electrophysiological research in rodents suggests that the hippocampus \cite{o1978hippocampus} and the entorhinal cortex \cite{moser2017spatial} are the neurological basis of cognitive maps. There, highly specialised neurons including place \cite{o1971hippocampus} and grid cells \cite{hafting2005microstructure} support map-like spatial codes, and thus enable spatial representation and navigation \cite{moser2008place}, and furthermore the construction of multi-scale maps \cite{geva2015spatial, kunz2019mesoscopic}. Also human fMRI studies during virtual navigation tasks have shown that the hippocampal and entorhinal spatial codes, together with areas in the frontal lobe, enable route planning during navigation \cite{spiers2006thoughts}, e.g. detours \cite{spiers2015solving}, shortcuts or efficient novel routes \cite{hartley2003well}, and in particular hierarchical spatial planning \cite{balaguer2016neural} based on distance preserving representations \cite{morgan2011distances}. 

Recent human fMRI studies even suggest that these map-like representations might not be restricted to physical space, i.e. places and spatial relations, but also extend to more abstract relations like in social and conceptual spaces \cite{epstein2017cognitive,park2021inferences,park2020map}, thereby contributing broadly to other cognitive domains \cite{schiller2015memory}, and thus enabling navigation and route planning in arbitrary abstract cognitive spaces \cite{bellmund2018navigating}. 

The hippocampus also plays a crucial role in episodic and declarative memory \cite{tulving1998episodic,reddy2021human}. Furthermore, the hippocampal formation, as a hub in brain connectivity \cite{battaglia2011hippocampus}, receives highly processed information via direct and indirect pathways from a large number of multi-modal areas of the cerebral cortex including language related areas in the frontal, temporal, and parietal lobe \cite{hickok2004dorsal}. Finally, some findings indicate that the hippocampus even contributes to the coding of narrative context \cite{milivojevic2016coding, morton2021concept}, and that memory representations, similar to the internal representation of space, systematically vary in scale along the hippocampal long axis \cite{collin2015memory}.
This scale might be used for goal directed navigation with different horizons \cite{brunec2019predictive} or even encode information from smaller episodes to more complex concepts  \cite{milivojevic2013mnemonic}. This geometry can also be modeled in artificial neural networks when performing an abstraction task \cite{bernardi2020geometry}. Cognitive maps therefore enable flexible planning through re-mapping of place cells and through the continuous (re-)scaling, generalization or detailed representation of information \cite{momennejad2020learning}. 

A number of computational models try to describe the hippocampal-entorhinal complex. For instance, the Tolman-Eichenbaum Machine describes hippocampal and entorhinal cell types and allows flexible transfer of structural knowledge \cite{whittington2020tolman}. Another framework that aims to describe the firing patterns of place cells in the hippocampus uses the successor representation (SR) as a building block for the construction of cognitive or predictive maps \cite{stachenfeld2014design, stachenfeld2017hippocampus}. The hierarchical structure in the entorhinal cortex can also be modeled by means of multi-scale successor representations \cite{MSSR}. Here, SR can be for example learned with a feature set of boundary vector cells \cite{de2019neurobiological} or with a sequence generation model inspired by the entorhinal-hippocampal circuit \cite{mcnamee2021flexible}.

To further investigate both the biological plausibility and potential machine learning applications of multi-scale SR and cognitive maps, we developed a neural network based simulation of place cell behavior under different circumstances. In particular, we trained a neural network to learn the SR for a simulated spatial environment and a navigation task in a virtual maze as proposed by Alvernhe et al. \cite{ratmaze}. In addition, we investigated if the applicability of our model extends from space to language as the hippocampus is known to also contribute to language processing \cite{piai_direct_2016}\cite{covington_expanding_2016}. Therefore, we created a model to learn a simplified artificial language. In particular, the model's task was to learn the underlying grammatical structure in terms of SR of words by observing exemplary input sentences only.

%=========================================================================
\section*{Methods}

\subsection*{Successor Representation}
The developed model is based on the principle of the successor representation (SR). As proposed by Stachenfeld et al. the SR can model the firing patterns of the place cells in the hippocampus \cite{stachenfeld2017hippocampus}. The SR was originally designed to build a representation of all possible future rewards $V(s)$ that may be achieved from each state $s$ within the state space over time \cite{SR_Original}. The future reward matrix $V(s)$ can be calculated for every state in the environment whereas the parameter $t$ indicates the number of time steps in the future that are taken into account, and $R(s_t)$ is the reward for state $s$ at time $t$. The discount factor $\gamma[0,1]$ reduces the relevance of states $s_t$ that are further in the future relative to the respective initial state $s_0$ (cf. Equation \ref{successor_representation}).  

\begin{equation}\label{successor_representation}
\centering
V(s) = E[\sum^{\infty}_{t=0}\gamma^t R(s_t)|s_0=s] 
\end{equation}
Here, $E[]$ denotes the expectation value.

The future reward matrix $V(s)$ can be re-factorized using the SR matrix $M$, which can be computed from the state transition probability matrix $T$ of successive states (cf. \ref{SR_refactor}). In case of supervised learning, the environments used for our model operate without specific rewards for each state. For the calculation of these SR we choose $R(s_t)=1$ for every state.

\begin{align}\label{SR_refactor}
    V(s) = \sum_{s'} M(s,s')R(s') && M = \sum^{\infty}_{t=0}\gamma^t T^t
\end{align}

\subsection*{Spatial Environment}
The spatial environments created in our framework are designed as discrete grid-like spatial rooms which can be freely explored by the agent. The neighboring states of a particular state are defined as direct successor states. Walls and barriers are not counted as possible successor state for a neighboring initial state. The squared room from the spatial exploration task consisted of 100 states arranged as a $10 \times 10$ rectangular grid (cf. Figure \ref{fig1}). The maze from the spatial navigation task was represented as a $15 \times 15$ rectangular grid, whereas only 94 states from all 225 states were "allowed" states that could be observed by the agent (cf. Figure \ref{fig3}).

\subsection*{Language Environment}
Additionally, we set up a state space with a non-spatial structure, i.e. a linguistic environment. The environment consists of 40 discrete states representing the vocabulary. Each state corresponds to a particular word, and each word belongs to one of the five different word classes: adjectives, verbs, nouns, pronouns and question words. The transition probabilities between subsequent words are defined according to a simplified syntax which consists of three types of linguistic constructions: an adjective-noun construction (cf. rule \ref{rule1}), a descriptive construction (cf. rule \ref{rule2}) and an interrogative construction (cf. rule \ref{rule3}).

\begin{subequations}\label{rules}
\begin{equation}\label{rule1}
    adjective \rightarrow noun
\end{equation}
\begin{equation}\label{rule2}
    pronoun \rightarrow verb \rightarrow adjective
\end{equation}
\begin{equation}\label{rule3}
    question \rightarrow pronoun \rightarrow verb
\end{equation}
\end{subequations}

The syntax rules, i.e. constructions, determine the transition probabilities for randomly chosen starting states and a word from the picked word group is set as label for the training data. The individual words from the successor word class are chosen with equal probability. The constructed sentences have therefore no particular meaning. For the language data set, 5,000 training samples were generated and the network was trained for 50 epochs.

\subsection*{Neural network architectures}
To be able to learn the SR by just observing the environment, we set up three-layered neural networks that learn the transition probabilities of the different environment. The input to the network is the momentary state encoded as one-hot vectors. Thus the number of neurons in the input layer is 100 in the exploration task, 225 in the navigation task and 45 in the linguistic task.

The hidden layer neurons have a ReLU activation function, whereas the number of neurons is equal to the corresponding number of input neurons in all networks. We also tested network architectures with a smaller number of hidden layer neurons (bottleneck). In all cases, the results where very similar.  

The softmax output layer gives a probability distribution for all successor states. Thus, the number of neurons in the output layer corresponds to the number of possible states and thus to the number of input neurons in all networks. 

The training data set is created by sampling trajectories through the spatial structure of the simulated environment. First, a random starting state is chosen as input and subsequently another random possible successor state is chosen from its neighbors as desired output. Walls are excluded as input states. For the first experiment we sampled 50,000 states and successor states, and trained for 10,000 epochs.

\subsubsection*{Reinforcement learning}
In an attampt to reproduce experimental data, we simulated a maze as proposed by Alvernhe et al. \cite{ratmaze}. Furthermore, a reward system is required for reinforcement learning (RL). Our RL approach enables us to define rewards in the spatial environments, which we use to simulate the food trays of the original experiment. The network structure is again a three-layered network, with a ReLU activation function for the hidden layer neurons and a softmax output layer, which yields the probabilities for the next actions. A DQN agent can choose from several actions depending on the number of the neighboring states belonging to the current state. If the agent chooses a wall state during training, the momentary training run is terminated and a new random starting state is chosen randomly. Training was performed for 10,000 epochs.

\subsubsection*{Transition probability and successor representation matrix}
After the training process, the network can predict all probabilities of successor states for any given initial state. Concatenating the predictions of all states leads to the transition probability (TP) matrix of our environments, which we use to calculate the SR matrix (cf. Equation \ref{SR_refactor}). In case of the supervised learning approach (spatial exploration task and language task), the output of the network is a vector shaped like a row of the respective environment's TP or SR matrix and can therefore directly be used to fill the TP or SR matrix, respectively. The reinforcement learning network however only yields the probabilities for the direct successors of a given state, which therefore need to be further extended to a vector containing all possible states of the environment.

\subsubsection*{Reproducing experimental grid cell firing patterns}
After training the network, the resulting SR matrices are evaluated. Therefore, each state encoded as one-hot vector is fed in as input to the network, and the resulting softmax output vectors are concatenated to built the SR matrix.   
The resulting SR matrix can be used to calculate their Eigendecompostion. The different Eigenvectors can be ordered according to their size, and are subsequently reshaped to fit the shape of the corresponding state space, i.e. simulated environment. The reshaped Eigenvectors are supposed to form grid like patterns, and to be a representation of the grid cells' receptive fields \cite{stachenfeld2017hippocampus}.

\subsection*{Multi-dimensional scaling}
A frequently used method to generate low-dimensional embeddings of high-dimensional data is t-distributed stochastic neighbor embedding (t-SNE) \cite{van2008visualizing}. However, in t-SNE the resulting low-dimensional projections can be highly dependent on the detailed parameter settings \cite{wattenberg2016use}, sensitive to noise, and may not preserve, but rather often scramble the global structure in data \cite{vallejos2019exploring, moon2019visualizing}.
In contrats, multi-Dimensional-Scaling (MDS) \cite{torgerson1952multidimensional, kruskal1964nonmetric,kruskal1978multidimensional,cox2008multidimensional} is an efficient embedding technique to visualize high-dimensional point clouds by projecting them onto a 2-dimensional plane. Furthermore, MDS has the decisive advantage that it is parameter-free and all mutual distances of the points are preserved, thereby conserving both the global and local structure of the underlying data. 

When interpreting patterns as points in high-dimensional space and dissimilarities between patterns as distances between corresponding points, MDS is an elegant method to visualize high-dimensional data. By color-coding each projected data point of a data set according to its label, the representation of the data can be visualized as a set of point clusters. For instance, MDS has already been applied to visualize for instance word class distributions of different linguistic corpora \cite{schilling2021analysis}, hidden layer representations (embeddings) of artificial neural networks \cite{schilling2021quantifying,krauss2021analysis}, structure and dynamics of recurrent neural networks \cite{krauss2019analysis, krauss2019recurrence, krauss2019weight}, or brain activity patterns assessed during e.g. pure tone or speech perception \cite{krauss2018statistical,schilling2021analysis}, or even during sleep \cite{krauss2018analysis,traxdorf2019microstructure}. 
In all these cases the apparent compactness and mutual overlap of the point clusters permits a qualitative assessment of how well the different classes separate.

\subsection*{Code Implementation}
The models were coded in Python. The neural networks were design using the Keras \cite{keras} and Keras-RL \cite{kerasrl} libraries. Mathematical operations were performed with numpy \cite{numpy} and scikit-learn \cite{scikit-learn} libraries.
Visualizations were realised with matplotlib \cite{matplot} and networkX \cite{networkX}.

%=========================================================================
\section*{Results}

\subsection*{Spatial environment}

\subsubsection*{Supervised learning reproduces basic firing patterns of place cells in rodents}
\begin{figure}[htbp]
    \centering
    \includegraphics[width = 13cm]{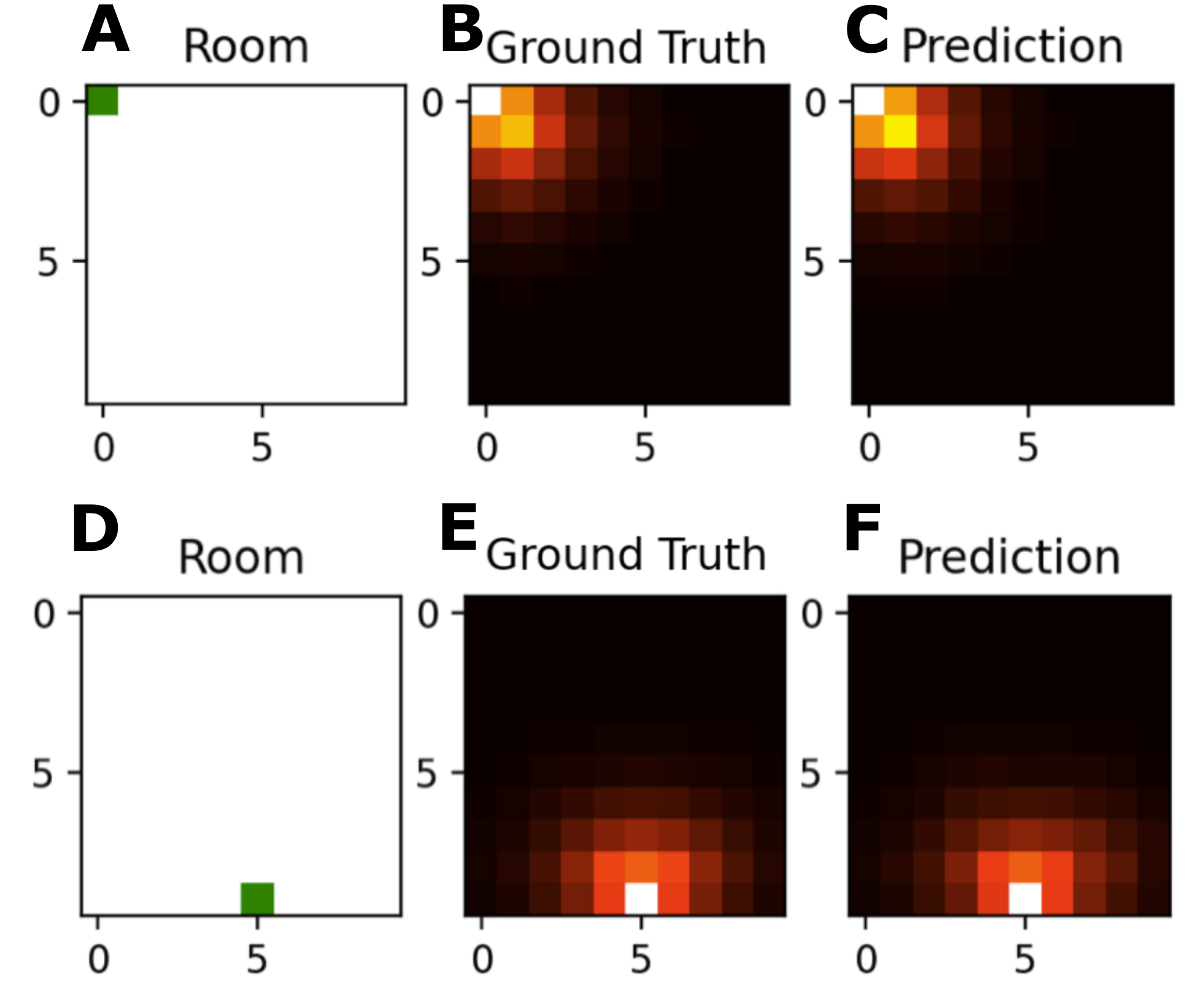}
    \caption{\textbf{Supervised learning to explore a spatial environment:} The SR for a 2D squared environment is learned with a supervised neural network. The small green squares indicate two sample starting positions (a, d). The corresponding SR are calculated and serve as ground truth (b, e). The neural network learns the transition probabilities for its direct neighbors and estimates the SR for a sequence length of t=10 (c, f).}
    \label{fig1}
\end{figure}

%Supervised Learning
In the supervised learning approach, the transition probabilities between neighboring places and hence the SR is learned by randomly observing places (states) and exploring their potential successors. In the simplest case of a 2D square environment without any obstacles, the transition probabilities from any starting place to all its eight neighbors are identical, i,e. uniformly distributed. Places at the walls (corners) of the room however only have five (three) neighboring states, so the transition probabilities corresponding to the remaining neighboring states representing those walls are zero, respectively. 
The resulting successor representations learned by the neural network are almost identical to the the ground truth (cf. Figure \ref{fig1}). Furthermore, these firing patterns are strikingly similar to those of place cells in rodents. They reflect the environment's spatial structure depending on obstacles and room shape \cite{krupic_grid_2015}, i.e. the intensity of the firing patterns is centered around the starting position (as this is also the most probable next state) and directed away from any walls into the open space (cf. Figure 1 \ref{fig1}), as described e.g. in \cite{PredictiveMap}, and found experimentally \cite{placecells}. 

\subsubsection*{Eigenvectors of learned SR resemble firing patterns of grid cells in rodents}

\begin{figure}[htbp]
    \centering
    \includegraphics[width = 15cm]{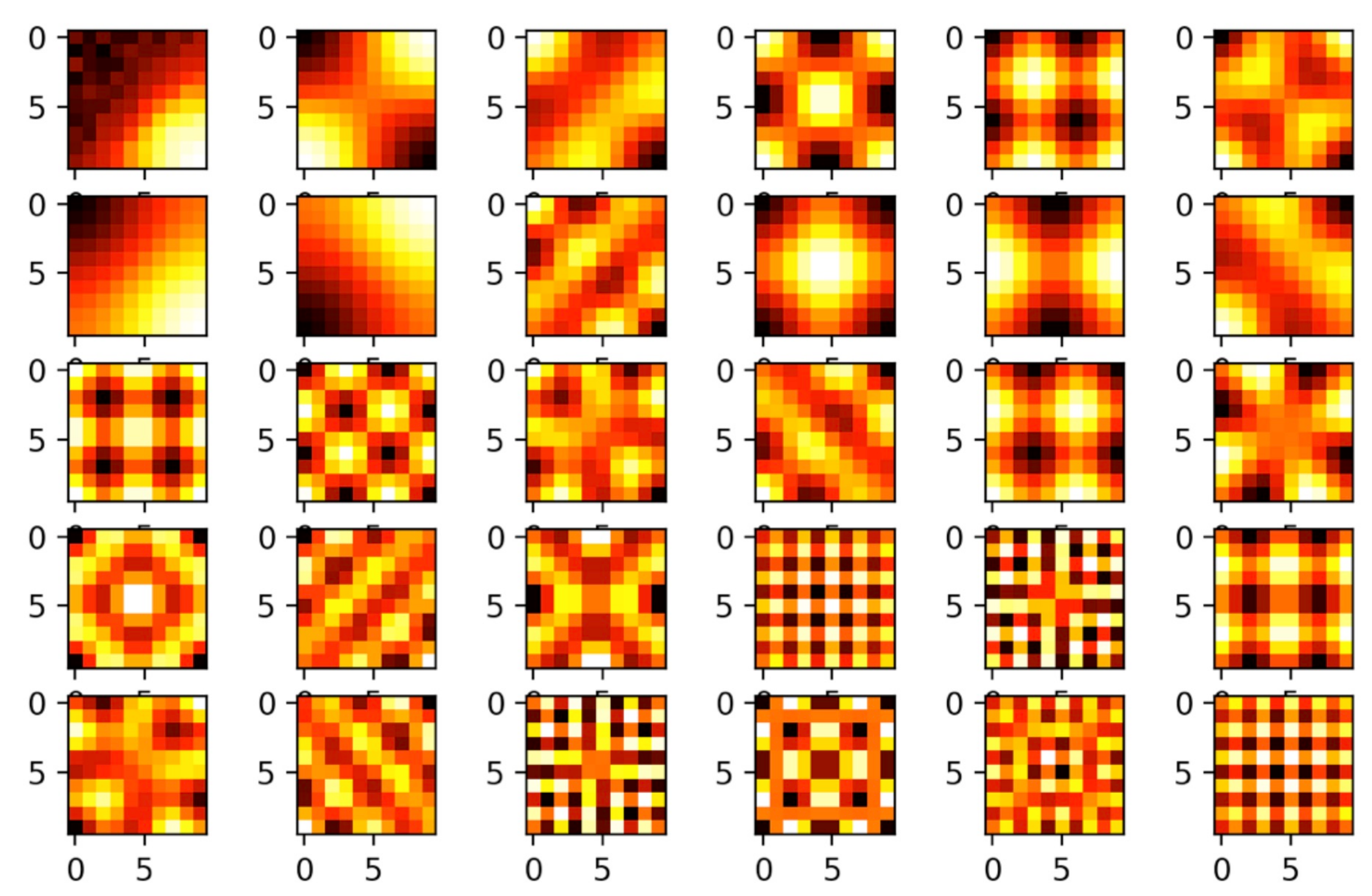}
    \caption{\textbf{Grid cell-like Eigenvectors of the SR matrix:} The firing patterns of grid cells in the entorhinal cortex are proposed to represent the Eigenvectors of the SR matrix \cite{stachenfeld2017hippocampus}. For the squared room depicted in Figure \ref{fig1}, the Eigenvectors for the first 30 Eigenvalues of the SR matrix are shown (re-shaped to the shape of the squared environment). Indeed, they resemble grid cell-like firing patterns. Furthermore, the grids vary in orientation and scaling, as observed in electrophysiological experiments in rodents.}
    \label{fig2}
\end{figure}

Stachenfeld et al. \cite{PredictiveMap} propose that the grid-like firing patterns of grid cells in the entorhinal cortex of rodents \cite{gridmeasure} may be explained by an Eigendecomposition of the SR matrix, whereas each individual grid cell would correspond to one Eigenvector. To test this assumption in the context of our framework, we calculated the Eigenvectors of the learned SR matrix as shown in Figure \ref{fig1}, and reshaped them to the shape of the environment. We find that this procedure actually leads to grid cell-like firing patterns (cf. Figure \ref{fig2}). The first 30 Eigenvectors ordered by increasing value of the corresponding Eigenvalues are shown in Figure \ref{fig2}. As known from neurobiology \cite{gridmeasure}, the grid-like patterns vary in orientation and mesh size (i.e. frequency). In particular, the smaller the corresponding Eigenvalue of the Eigenvector, the smaller the mesh size, i.e. more fine-grained the resulting grid, becomes (cf. Figure \ref{fig2}). Furthermore it is known that, the individual orientation of the grid cells' firing patterns follows no particular order, whereas the mesh size of the grids varies systematically along the long axis of the entorhinal cortex \cite{brun2008progressive}. This feature, especially, is thought to enable multi-scale mapping, route planing and navigation \cite{MSSR}.

\subsubsection*{Reinforcement learning reproduces basic firing patterns of place cells in rodents}

\begin{figure}[htbp]
    \centering
    \includegraphics[width = 15cm]{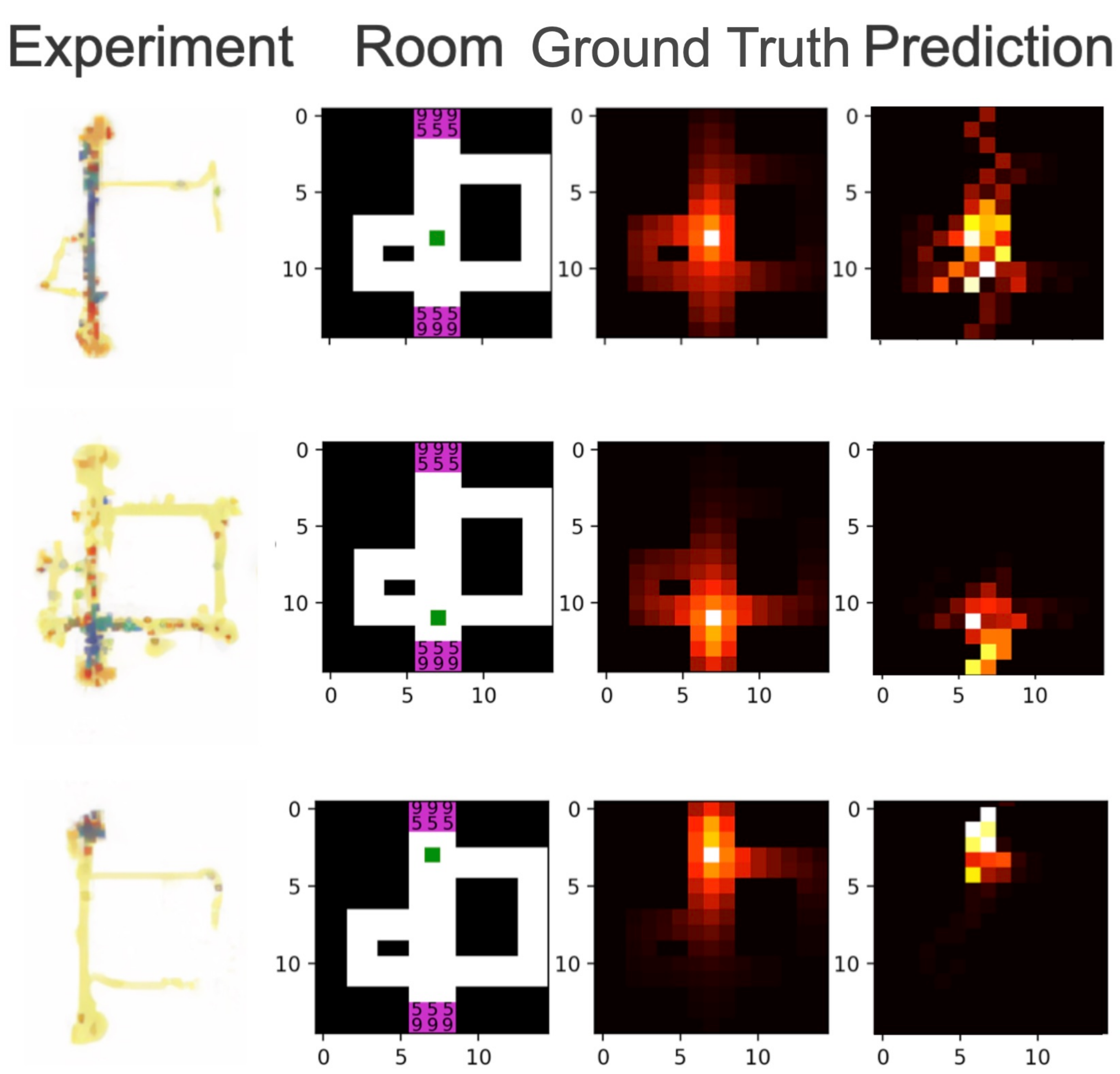}
    \caption{\textbf{Reinforcement learning to navigate a spatial environment:} We reproduced the rat maze experiment published by Alvernhe et al. \cite{ratmaze} (left column). Therefore, we simulated the corresponding maze environment, small green squares indicate three different sample starting positions (second column). Based on the transition probabilities to neighboring states we calculated the successor representation of the maze as ground truth (third column). The predicted SR of the trained network, i.e. the firing patterns of the artificial place cells are very similar to the underlying ground truth (right column).}
    \label{fig3}
\end{figure}

%Reinforcement Learning
In contrast to a goal-free random walk in order to explore a novel environment (as in the previous setting), navigation is usually driven by a specific goal or reward \cite{gerum2020sparsity} like, e.g. food. The task is therefore ideally suited for reinforcement learning (RL) \cite{sutton2018reinforcement}. In our simulation, we reproduced a classical rodent maze experiment presented by Alvernhe et al. \cite{ratmaze}. As in the supervised learning setting, the successor representations learned in the RL setting are very similar to the ground truth (cf. Figure \ref{fig3}). Also, the resulting place cell firing patterns closely resemble those of place cells in rodents during maze navigation tasks.
The SR place fields are clearly different from those obtained without any reward in the goal-free exploration task (cf. Figure \ref{fig1}). A position close to a reward state is associated with highly localized firing patterns, whereas the highest successor probabilities are directed towards the reward states. Furthermore, places in the middle of the maze are associated with firing patterns that are stretched parallel to the orientation of the maze's main corridor. Highest successor probabilities are localized around the starting position, but still also in reach of the reward states. The side arms of the maze are a detour to the goal (reward position), and are therefore mainly ignored by the network, i.e. associated with the lowest successor state probabilities.

%Language Model
\subsection*{Linguistic structures}

\subsubsection*{Linguistic constructions define a network-like linguistic map}

\begin{figure}[htbp]
    \centering
    \includegraphics[width = 20cm]{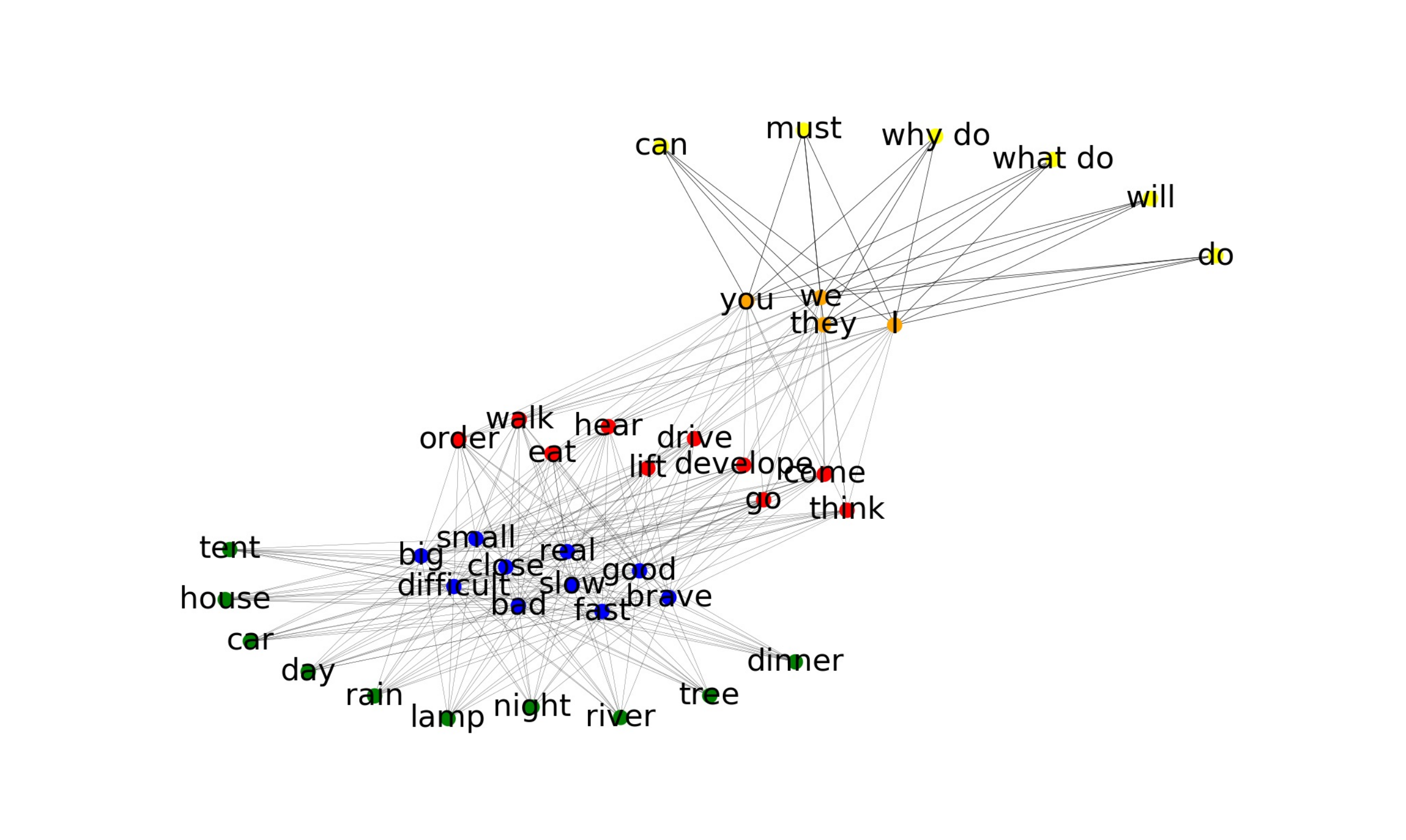}
    \caption{\textbf{Network-like map of linguistic constructions:} The simplified language model consists of five different word classes and three linguistic constructions defining allowed word class transitions. The word transition matrix can be visualized as a graph or network-like map, whereas each word corresponds to a node, and edges represent possible word transitions. Different node colors indicate different word classes. Note that, edges for transition probabilities smaller than $10^{-4}$ are not shown for better readability. }
    \label{fig4}
\end{figure}

Cognitive maps are however not restricted to physical space. On the contrary, cognitive maps may also be applied to arbitrary abstract and complex state spaces. In general, any state space can be represented as a graph. In this case, nodes correspond to states, and edges to state transitions. A prime example of such graph-like (or network-like) state space representations is language. In cognitive linguistics, there is an overall agreement on the fact, that language is represented as a network in the human mind \cite{goldberg1995constructions,goldberg2003constructions,goldberg2006constructions,goldberg2019explain,diessel2019usage,diessel2020dynamic,levshina2021grammar}, whereas the nodes correspond to linguistic units at different hierarchical levels from phonemes, through words, to idioms and abstract argument structure constructions \cite{goldberg1995constructions}. In particular, "the nodes at one level of analysis are networks at another level of analysis" \cite{diessel2019grammar}. Hence, multi-scale SR \cite{MSSR} appears to be an ideal theoretical framework to explain language representation and processing in the human mind, whereas the systematically varying grid-scale along the long axis of the entorhinal cortex \cite{brun2008progressive} might explain its implementation in the human brain.
To investigate this hypothesis, we constructed as a first step a simplified language as described in detail in the Methods section. The lexicon together with the three linguistic constructions result in a network-like linguistic map (cf. Figure \ref{fig4}), that has to be learned by the neural network.

\subsubsection*{The neural network learns state TP and SR matrices}

\begin{figure}[h]
    \centering
    \includegraphics[width = 19cm]{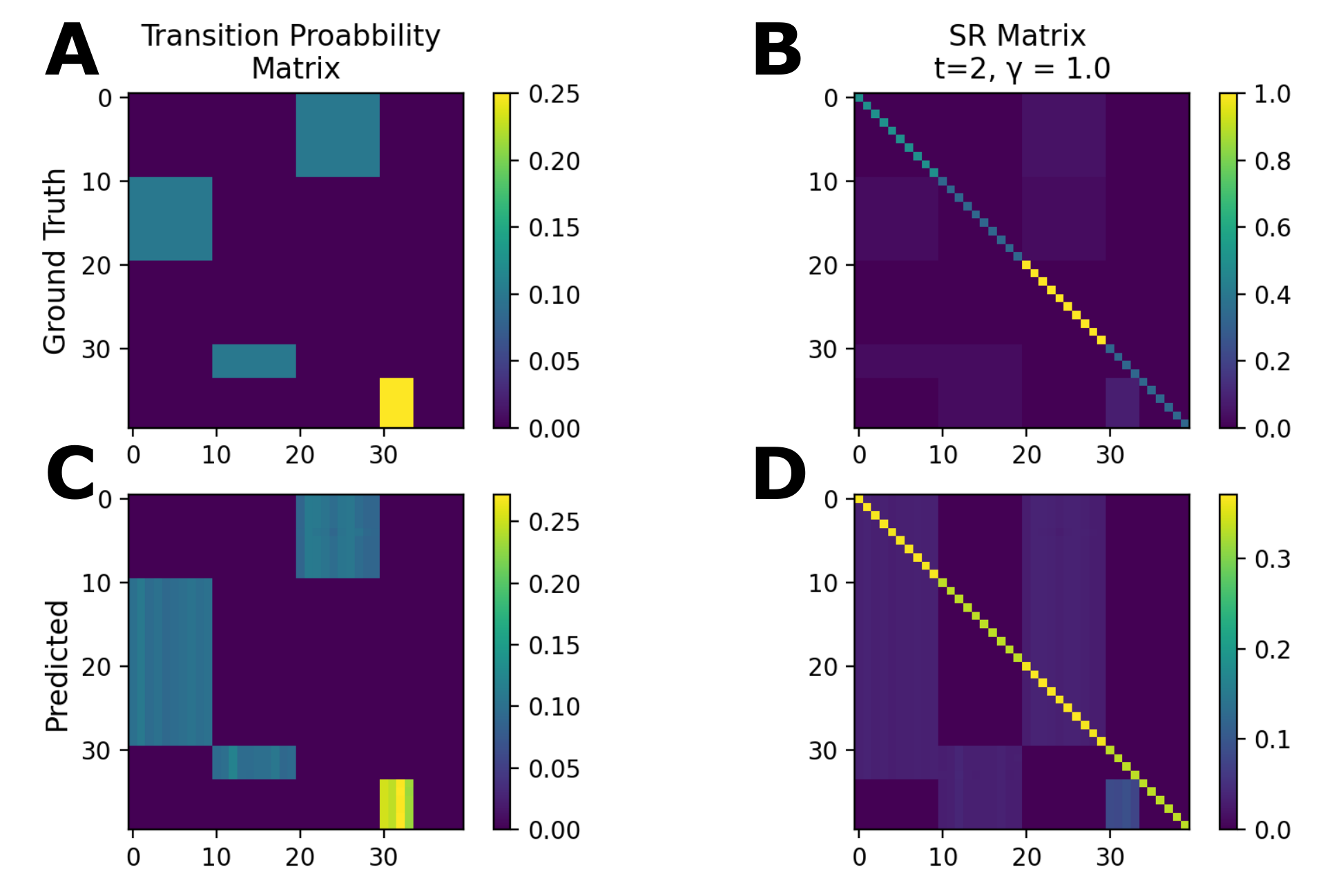}
    \caption{\textbf{Word transition probability and word successor representation  matrices:} After training on the linguistic data, the neural network predicts the word transition probabilities (TP) and the word successor representations (SR). As ground truth, the calculated TP matrix (a) and the SR matrix (b) for $t=2$ and $\gamma=1$ are shown. The corresponding predictions learned by the network are very similar to the ground truth in both cases (c, d). States (0-39) correspond to words. 0-9: adjectives, 10-19: verbs, 20-29: nouns, 30-34: pronouns, 35-40: question words.}
    \label{fig5}
\end{figure}

The learned behaviour of the network in the state space can be displayed as a state TP or SR matrix. After training, the TP matrix which is predicted by the network is very similar to the ground truth (cf. Figure \ref{fig5}a,c). However the network also predicts adjectives (states 0-10) as successors of nouns (states 20-30), even though this transition does not explicitly exist in the three pre-defined constructions. Consequently, the network's SR matrix is also slightly different from the ground truth (cf. Figure \ref{fig5}b,d).

\subsubsection*{Word classes spontaneously emerge as clusters in the TP and SR vector space}

\begin{figure}[htbp]
    \centering
    \includegraphics[width = 15cm]{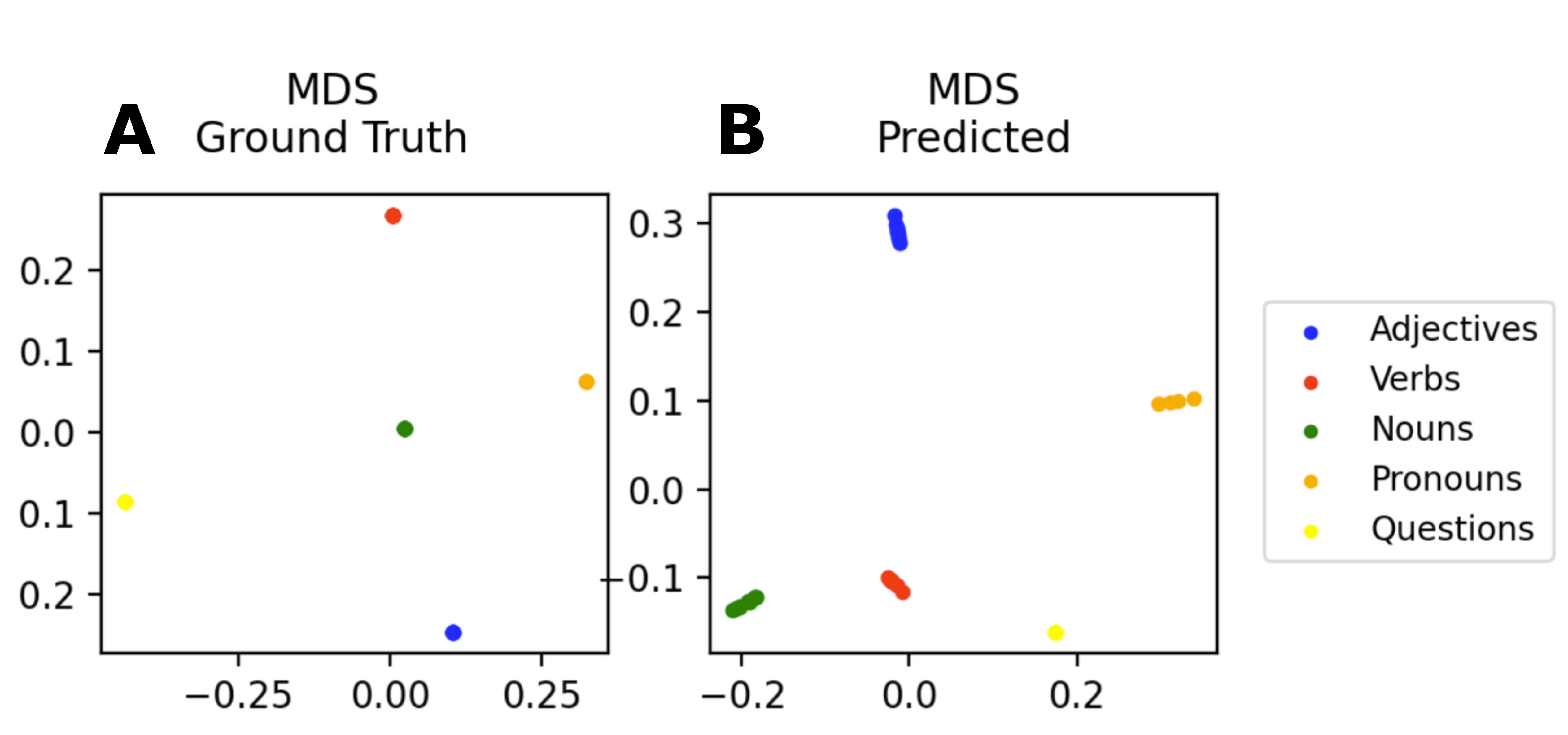}
    \caption{\textbf{MDS of the word transition probability vectors:} A two-dimensional projection of the 40-dimensional word TP vectors (rows in TP matrix) for the calculated ground truth (a) and the learned TP matrix (b). In both cases, the words build clearly separated, dense clusters according to the respective word class. Different colors correspond to word classes. Note that, scaling of the axes is in arbitrary units since coordinates have no particular meaning other than indicating the relative positions of the projected vectors.}
    \label{fig6}
\end{figure}

\begin{figure}[htbp]
    \centering
    \includegraphics[width = 15cm]{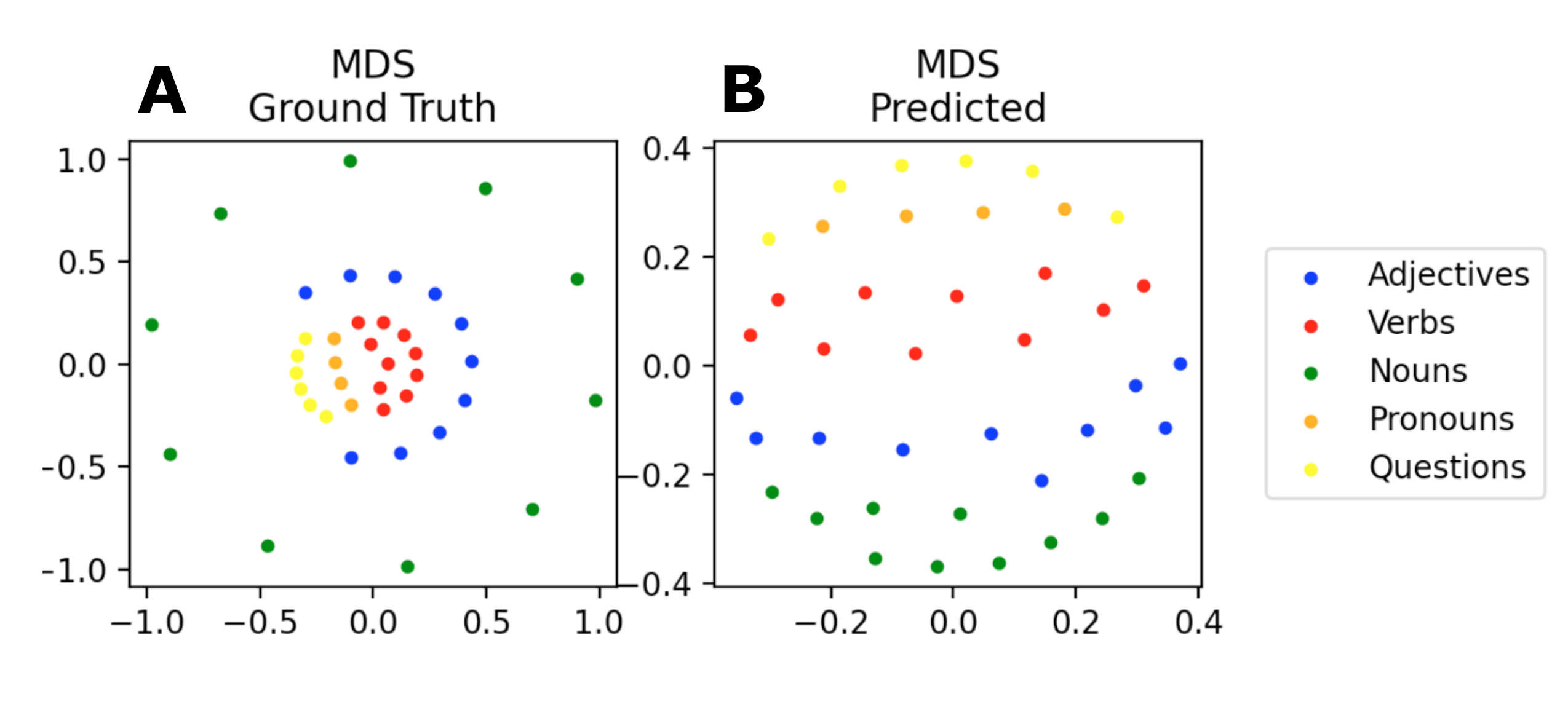}
    \caption{\textbf{MDS of the word successor representation vectors:} A two-dimensional projection of the 40-dimensional word SR vectors (rows in SR matrix) for the calculated ground truth (a) and the learned SR matrix (b). In both cases, the words cluster according to the respective word class. Different colors correspond to word classes again. However, the resulting clusters are less dense and located closer to each other than for the TP vectors. Since SR vectors cover several time steps, whereas TP vectors only cover a single time step in the future, this result is intuitive. Note that, scaling of the axes is in arbitrary units since coordinates have no particular meaning other than indicating the relative positions of the projected vectors.}
    \label{fig7}
\end{figure}

The transition probabilities from a given word to all other 40 words (rows in the TP matrix), as well as the corresponding successor probabilities (rows in the SR matrix) can be represented as vectors, and hence may be interpreted as points in a 40-dimensional TP or SR space respectively, whereas each word corresponds to a particular point. In order to further investigate the properties of these high-dimensional representations, we visualize both the TP and the SR space using multi-dimensional scaling. In particular, the 40-dimensional TP and SR vector representations of each word are projected onto a two-dimensional plane as described in detail in the Methods section. By color-coding each word according to its word class, we observed putative clustering of the vocabulary. Remarkably, the words actually cluster according to their word classes (cf. Figures \ref{fig6} and \ref{fig7}), even though, this information was not provided (e.g. as an additional label for each word) to the neural network at any time during training.

% t=0 SR matrix = trans prob matrix

% t=2 SR matrix

%===========================================
\section*{Discussion}

In this study, we demonstrated that efficient successor representations can be learned by artificial neural networks in different scenarios. The emerging representations share important properties with network-like cognitive maps, enabling e.g. navigation in arbitrary abstract and conceptual spaces, and thereby broadly supporting domain-general cognition, as proposed by Bellmund et al. \cite{bellmund2018navigating}.

In particular, we created a model, which can learn the SR for spatial and non-spatial environments. The model successfully reproduced experimentally observed firing patterns of place and grid cells in simulated spatial environments in two different scenarios. First, an exploration task based on supervised learning in a squared room without any obstacles, and second, a navigation task based on reinforcement learning in a simulated maze. Furthermore our neural network model learned the underlying word classes of a simplified artificial language framework just by observing sequences of words.

The involvement of the entorhinal-hippocampal complex -- as being the most probable candidate structure underlying network-like cognitive maps and multi-scale navigation \cite{PredictiveMap, park2020map, park2021inferences, momennejad2020learning, MSSR} -- in language processing has already been experimentally demonstrated \cite{piai_direct_2016, covington_expanding_2016}. Our study further supports, in particular, the involvement of place cells, as being the nodes of the "language network" as suggested in cognitive linguistics \cite{goldberg1995constructions,goldberg2003constructions,goldberg2006constructions,goldberg2019explain,diessel2019usage,diessel2020dynamic,levshina2021grammar}. Early language acquisition, especially, is driven by passive listening \cite{kuhl_brain_2010} and implicit learning \cite{dabrowska2014implicit}. Our model replicates learning by listening and therefore resembles a realistic scenario.

The varying grid cell scaling along the long axis of the entorhinal cortex is known to be associated with hierarchical memory content  \cite{collin2015memory}. The Eigenvectors of the SR matrix are strikingly similar to the firing patterns of grid cells, and therefore provide a putative explanation of the computational mechanisms underlying grid cell coding. These multi-scale representations are perfectly suited to map hierarchical linguistic structures from phonemes through words to sentences, and even beyond, like e.g. events or entire narratives. Indeed, recent neuroimaging studies provide evidence for the existence of "event nodes" in the human hippocampus \cite{milivojevic2016coding}.

Since our neural network model is able learn the underlying structure of a simplified language, we speculate that also the human hippocampal-entorhinal complex similarly encodes the complex linguistic structures of the languages learned by a given individual. Therefore learning further languages of similar structure as previously learned languages might be easier due to the fact that the multi-scale representations and cognitive maps in the hippocampus can be more easily transferred and re-mapped \cite{park2020map, park2021inferences}, i.e. re-used, in major parts to the new language.

Whether the hippocampus is actually involved in multi-scale representation and processing of linguistic structures across several hierarchies needs to be verified experimentally and theoretically. Neuroimaging studies during natural language perception and production, like for instance listening to audiobooks \cite{schilling2021analysis}, need to be performed. Only continuous, connected speech and language provides such corpus-like rich linguistic structures, being crucial to assess putative multi-scale processing. Additionally, further theoretical studies are needed to extend the presented model, and to apply it to more complex and naturalistic linguistic tasks, like e.g. word prediction in a natural language scenario.

As recently suggested, the neuroscience of spatial navigation might be of particular importance for artificial intelligence research \cite{bermudez2020neuroscience}. A neural network implementation of hippocampal successor representations, especially, promises advances in both fields. Following the research agenda of Cognitive Computational Neuroscience proposed by Kriegeskorte et al. \cite{kriegeskorte2018cognitive}, neuroscience and cognitive science benefit from such models by gaining deeper understanding of brain computations \cite{schilling2020intrinsic, krauss2018cross, krauss2021analysis}. Conversely, for artificial intelligence and machine learning, neural network-based multi-scale successor representations to learn and process structural knowledge (as an example of neuroscience-inspired artificial intelligence \cite{hassabis2017neuroscience}), might be a further step to overcome the limitations of contemporary deep learning \cite{marcus2018deep, yang2021neural, gerum2021integration, maier2019learning} and towards human-level artificial general intelligence.

%=========================================================================
\FloatBarrier
\section*{Acknowledgments}

This work was funded by the Deutsche Forschungsgemeinschaft (DFG, German Research Foundation): grant KR5148/2-1 to PK -- project number 436456810.

%=========================================================================
\section*{Author contributions}
PS and CS performed computer simulations. PK and AM designed and supervised the study. All authors discussed the results and wrote the manuscript.

%=========================================================================
\section*{Competing interests}
The authors declare no competing financial interests.

%=========================================================================
\FloatBarrier

%\bibliographystyle{unsrt}
%\bibliography{literature}

\begin{thebibliography}{10}

\bibitem{tolman1948cognitive}
Edward~C Tolman.
\newblock Cognitive maps in rats and men.
\newblock {\em Psychological review}, 55(4):189, 1948.

\bibitem{o1978hippocampus}
John O'keefe and Lynn Nadel.
\newblock {\em The hippocampus as a cognitive map}.
\newblock Oxford university press, 1978.

\bibitem{moser2017spatial}
Edvard~I Moser, May-Britt Moser, and Bruce~L McNaughton.
\newblock Spatial representation in the hippocampal formation: a history.
\newblock {\em Nature neuroscience}, 20(11):1448--1464, 2017.

\bibitem{o1971hippocampus}
John O'Keefe and Jonathan Dostrovsky.
\newblock The hippocampus as a spatial map: preliminary evidence from unit
  activity in the freely-moving rat.
\newblock {\em Brain research}, 1971.

\bibitem{hafting2005microstructure}
Torkel Hafting, Marianne Fyhn, Sturla Molden, May-Britt Moser, and Edvard~I
  Moser.
\newblock Microstructure of a spatial map in the entorhinal cortex.
\newblock {\em Nature}, 436(7052):801--806, 2005.

\bibitem{moser2008place}
Edvard~I Moser, Emilio Kropff, and May-Britt Moser.
\newblock Place cells, grid cells, and the brain's spatial representation
  system.
\newblock {\em Annu. Rev. Neurosci.}, 31:69--89, 2008.

\bibitem{geva2015spatial}
Maya Geva-Sagiv, Liora Las, Yossi Yovel, and Nachum Ulanovsky.
\newblock Spatial cognition in bats and rats: from sensory acquisition to
  multiscale maps and navigation.
\newblock {\em Nature Reviews Neuroscience}, 16(2):94--108, 2015.

\bibitem{kunz2019mesoscopic}
Lukas Kunz, Shachar Maidenbaum, Dong Chen, Liang Wang, Joshua Jacobs, and
  Nikolai Axmacher.
\newblock Mesoscopic neural representations in spatial navigation.
\newblock {\em Trends in cognitive sciences}, 23(7):615--630, 2019.

\bibitem{spiers2006thoughts}
Hugo~J Spiers and Eleanor~A Maguire.
\newblock Thoughts, behaviour, and brain dynamics during navigation in the real
  world.
\newblock {\em Neuroimage}, 31(4):1826--1840, 2006.

\bibitem{spiers2015solving}
Hugo~J Spiers and Sam~J Gilbert.
\newblock Solving the detour problem in navigation: a model of prefrontal and
  hippocampal interactions.
\newblock {\em Frontiers in human neuroscience}, 9:125, 2015.

\bibitem{hartley2003well}
Tom Hartley, Eleanor~A Maguire, Hugo~J Spiers, and Neil Burgess.
\newblock The well-worn route and the path less traveled: distinct neural bases
  of route following and wayfinding in humans.
\newblock {\em Neuron}, 37(5):877--888, 2003.

\bibitem{balaguer2016neural}
Jan Balaguer, Hugo Spiers, Demis Hassabis, and Christopher Summerfield.
\newblock Neural mechanisms of hierarchical planning in a virtual subway
  network.
\newblock {\em Neuron}, 90(4):893--903, 2016.

\bibitem{morgan2011distances}
Lindsay~K Morgan, Sean~P MacEvoy, Geoffrey~K Aguirre, and Russell~A Epstein.
\newblock Distances between real-world locations are represented in the human
  hippocampus.
\newblock {\em Journal of Neuroscience}, 31(4):1238--1245, 2011.

\bibitem{epstein2017cognitive}
Russell~A Epstein, Eva~Zita Patai, Joshua~B Julian, and Hugo~J Spiers.
\newblock The cognitive map in humans: spatial navigation and beyond.
\newblock {\em Nature neuroscience}, 20(11):1504--1513, 2017.

\bibitem{park2021inferences}
Seongmin~A Park, Douglas~S Miller, and Erie~D Boorman.
\newblock Inferences on a multidimensional social hierarchy use a grid-like
  code.
\newblock {\em bioRxiv}, pages 2020--05, 2021.

\bibitem{park2020map}
Seongmin~A Park, Douglas~S Miller, Hamed Nili, Charan Ranganath, and Erie~D
  Boorman.
\newblock Map making: Constructing, combining, and inferring on abstract
  cognitive maps.
\newblock {\em BioRxiv}, page 810051, 2020.

\bibitem{schiller2015memory}
Daniela Schiller, Howard Eichenbaum, Elizabeth~A Buffalo, Lila Davachi, David~J
  Foster, Stefan Leutgeb, and Charan Ranganath.
\newblock Memory and space: towards an understanding of the cognitive map.
\newblock {\em Journal of Neuroscience}, 35(41):13904--13911, 2015.

\bibitem{bellmund2018navigating}
Jacob~LS Bellmund, Peter G{\"a}rdenfors, Edvard~I Moser, and Christian~F
  Doeller.
\newblock Navigating cognition: Spatial codes for human thinking.
\newblock {\em Science}, 362(6415), 2018.

\bibitem{tulving1998episodic}
Endel Tulving and Hans~J Markowitsch.
\newblock Episodic and declarative memory: role of the hippocampus.
\newblock {\em Hippocampus}, 8(3):198--204, 1998.

\bibitem{reddy2021human}
Leila Reddy, Benedikt Zoefel, Jessy~K Possel, Judith Peters, Doris~E
  Dijksterhuis, Marlene Poncet, Elisabeth~CW van Straaten, Johannes~C Baayen,
  Sander Idema, and Matthew~W Self.
\newblock Human hippocampal neurons track moments in a sequence of events.
\newblock {\em Journal of Neuroscience}, 41(31):6714--6725, 2021.

\bibitem{battaglia2011hippocampus}
Francesco~P Battaglia, Karim Benchenane, Anton Sirota, Cyriel~MA Pennartz, and
  Sidney~I Wiener.
\newblock The hippocampus: hub of brain network communication for memory.
\newblock {\em Trends in cognitive sciences}, 15(7):310--318, 2011.

\bibitem{hickok2004dorsal}
Gregory Hickok and David Poeppel.
\newblock Dorsal and ventral streams: a framework for understanding aspects of
  the functional anatomy of language.
\newblock {\em Cognition}, 92(1-2):67--99, 2004.

\bibitem{milivojevic2016coding}
Branka Milivojevic, Meryl Varadinov, Alejandro~Vicente Grabovetsky, Silvy~HP
  Collin, and Christian~F Doeller.
\newblock Coding of event nodes and narrative context in the hippocampus.
\newblock {\em Journal of Neuroscience}, 36(49):12412--12424, 2016.

\bibitem{morton2021concept}
Neal~W Morton and Alison~R Preston.
\newblock Concept formation as a computational cognitive process.
\newblock {\em Current Opinion in Behavioral Sciences}, 38:83--89, 2021.

\bibitem{collin2015memory}
Silvy~HP Collin, Branka Milivojevic, and Christian~F Doeller.
\newblock Memory hierarchies map onto the hippocampal long axis in humans.
\newblock {\em Nature neuroscience}, 18(11):1562--1564, 2015.

\bibitem{brunec2019predictive}
Iva~K Brunec and Ida Momennejad.
\newblock Predictive representations in hippocampal and prefrontal hierarchies.
\newblock {\em bioRxiv}, page 786434, 2019.

\bibitem{milivojevic2013mnemonic}
Branka Milivojevic and Christian~F Doeller.
\newblock Mnemonic networks in the hippocampal formation: From spatial maps to
  temporal and conceptual codes.
\newblock {\em Journal of Experimental Psychology: General}, 142(4):1231, 2013.

\bibitem{bernardi2020geometry}
Silvia Bernardi, Marcus~K Benna, Mattia Rigotti, J{\'e}r{\^o}me Munuera,
  Stefano Fusi, and C~Daniel Salzman.
\newblock The geometry of abstraction in the hippocampus and prefrontal cortex.
\newblock {\em Cell}, 183(4):954--967, 2020.

\bibitem{momennejad2020learning}
Ida Momennejad.
\newblock Learning structures: Predictive representations, replay, and
  generalization.
\newblock {\em Current Opinion in Behavioral Sciences}, 32:155--166, 2020.

\bibitem{whittington2020tolman}
James~CR Whittington, Timothy~H Muller, Shirley Mark, Guifen Chen, Caswell
  Barry, Neil Burgess, and Timothy~EJ Behrens.
\newblock The tolman-eichenbaum machine: Unifying space and relational memory
  through generalization in the hippocampal formation.
\newblock {\em Cell}, 183(5):1249--1263, 2020.

\bibitem{stachenfeld2014design}
Kimberly~L Stachenfeld, Matthew Botvinick, and Samuel~J Gershman.
\newblock Design principles of the hippocampal cognitive map.
\newblock {\em Advances in neural information processing systems},
  27:2528--2536, 2014.

\bibitem{stachenfeld2017hippocampus}
Kimberly~L Stachenfeld, Matthew~M Botvinick, and Samuel~J Gershman.
\newblock The hippocampus as a predictive map.
\newblock {\em Nature neuroscience}, 20(11):1643, 2017.

\bibitem{MSSR}
Ida Momennejad and Marc~W. Howard.
\newblock Predicting the future with multi-scale successor representations.
\newblock {\em bioRxiv}, 2018.

\bibitem{de2019neurobiological}
William De~Cothi and Caswell Barry.
\newblock Neurobiological successor features for spatial navigation.
\newblock {\em BioRxiv}, page 789412, 2019.

\bibitem{mcnamee2021flexible}
Daniel~C McNamee, Kimberly~L Stachenfeld, Matthew~M Botvinick, and Samuel~J
  Gershman.
\newblock Flexible modulation of sequence generation in the
  entorhinal--hippocampal system.
\newblock {\em Nature Neuroscience}, 24(6):851--862, 2021.

\bibitem{ratmaze}
Alice Alvernhe, Etienne Save, and Bruno Poucet.
\newblock Local remapping of place cell firing in the tolman detour task.
\newblock {\em The European journal of neuroscience}, 33:1696--705, 03 2011.

\bibitem{piai_direct_2016}
Vitória Piai, Kristopher~L. Anderson, Jack~J. Lin, Callum Dewar, Josef
  Parvizi, Nina~F. Dronkers, and Robert~T. Knight.
\newblock Direct brain recordings reveal hippocampal rhythm underpinnings of
  language processing.
\newblock {\em Proceedings of the National Academy of Sciences of the United
  States of America}, 113:11366--11371, 2016.

\bibitem{covington_expanding_2016}
Natalie~V. Covington and Melissa~C. Duff.
\newblock Expanding the language network: {Direct} contributions from the
  hippocampus.
\newblock {\em Trends in cognitive sciences}, 20:869--870, December 2016.

\bibitem{SR_Original}
Peter Dayan.
\newblock {Improving Generalization for Temporal Difference Learning: The
  Successor Representation}.
\newblock {\em Neural Computation}, 5(4):613--624, 07 1993.

\bibitem{van2008visualizing}
Laurens Van~der Maaten and Geoffrey Hinton.
\newblock Visualizing data using t-sne.
\newblock {\em Journal of machine learning research}, 9(11), 2008.

\bibitem{wattenberg2016use}
Martin Wattenberg, Fernanda Vi{\'e}gas, and Ian Johnson.
\newblock How to use t-sne effectively.
\newblock {\em Distill}, 1(10):e2, 2016.

\bibitem{vallejos2019exploring}
Catalina~A Vallejos.
\newblock Exploring a world of a thousand dimensions.
\newblock {\em Nature biotechnology}, 37(12):1423--1424, 2019.

\bibitem{moon2019visualizing}
Kevin~R Moon, David van Dijk, Zheng Wang, Scott Gigante, Daniel~B Burkhardt,
  William~S Chen, Kristina Yim, Antonia van~den Elzen, Matthew~J Hirn, Ronald~R
  Coifman, et~al.
\newblock Visualizing structure and transitions in high-dimensional biological
  data.
\newblock {\em Nature biotechnology}, 37(12):1482--1492, 2019.

\bibitem{torgerson1952multidimensional}
Warren~S Torgerson.
\newblock Multidimensional scaling: I. theory and method.
\newblock {\em Psychometrika}, 17(4):401--419, 1952.

\bibitem{kruskal1964nonmetric}
Joseph~B Kruskal.
\newblock Nonmetric multidimensional scaling: a numerical method.
\newblock {\em Psychometrika}, 29(2):115--129, 1964.

\bibitem{kruskal1978multidimensional}
Joseph~B Kruskal.
\newblock {\em Multidimensional scaling}.
\newblock Number~11. Sage, 1978.

\bibitem{cox2008multidimensional}
Michael~AA Cox and Trevor~F Cox.
\newblock Multidimensional scaling.
\newblock In {\em Handbook of data visualization}, pages 315--347. Springer,
  2008.

\bibitem{schilling2021analysis}
Achim Schilling, Rosario Tomasello, Malte~R Henningsen-Schomers, Alexandra
  Zankl, Kishore Surendra, Martin Haller, Valerie Karl, Peter Uhrig, Andreas
  Maier, and Patrick Krauss.
\newblock Analysis of continuous neuronal activity evoked by natural speech
  with computational corpus linguistics methods.
\newblock {\em Language, Cognition and Neuroscience}, 36(2):167--186, 2021.

\bibitem{schilling2021quantifying}
Achim Schilling, Andreas Maier, Richard Gerum, Claus Metzner, and Patrick
  Krauss.
\newblock Quantifying the separability of data classes in neural networks.
\newblock {\em Neural Networks}, 139:278--293, 2021.

\bibitem{krauss2021analysis}
Patrick Krauss, Claus Metzner, Nidhi Joshi, Holger Schulze, Maximilian
  Traxdorf, Andreas Maier, and Achim Schilling.
\newblock Analysis and visualization of sleep stages based on deep neural
  networks.
\newblock {\em Neurobiology of sleep and circadian rhythms}, 10:100064, 2021.

\bibitem{krauss2019analysis}
Patrick Krauss, Alexandra Zankl, Achim Schilling, Holger Schulze, and Claus
  Metzner.
\newblock Analysis of structure and dynamics in three-neuron motifs.
\newblock {\em Frontiers in Computational Neuroscience}, 13:5, 2019.

\bibitem{krauss2019recurrence}
Patrick Krauss, Karin Prebeck, Achim Schilling, and Claus Metzner.
\newblock Recurrence resonance” in three-neuron motifs.
\newblock {\em Frontiers in computational neuroscience}, page~64, 2019.

\bibitem{krauss2019weight}
Patrick Krauss, Marc Schuster, Verena Dietrich, Achim Schilling, Holger
  Schulze, and Claus Metzner.
\newblock Weight statistics controls dynamics in recurrent neural networks.
\newblock {\em PloS one}, 14(4):e0214541, 2019.

\bibitem{krauss2018statistical}
Patrick Krauss, Claus Metzner, Achim Schilling, Konstantin Tziridis, Maximilian
  Traxdorf, Andreas Wollbrink, Stefan Rampp, Christo Pantev, and Holger
  Schulze.
\newblock A statistical method for analyzing and comparing spatiotemporal
  cortical activation patterns.
\newblock {\em Scientific reports}, 8(1):1--9, 2018.

\bibitem{krauss2018analysis}
Patrick Krauss, Achim Schilling, Judith Bauer, Konstantin Tziridis, Claus
  Metzner, Holger Schulze, and Maximilian Traxdorf.
\newblock Analysis of multichannel eeg patterns during human sleep: a novel
  approach.
\newblock {\em Frontiers in human neuroscience}, 12:121, 2018.

\bibitem{traxdorf2019microstructure}
Maximilian Traxdorf, Patrick Krauss, Achim Schilling, Holger Schulze, and
  Konstantin Tziridis.
\newblock Microstructure of cortical activity during sleep reflects respiratory
  events and state of daytime vigilance.
\newblock {\em Somnologie}, 23(2):72--79, 2019.

\bibitem{keras}
Fran\c{c}ois Chollet et~al.
\newblock Keras, 2015.

\bibitem{kerasrl}
Matthias Plappert.
\newblock keras-rl, 2016.

\bibitem{numpy}
Charles~R. Harris, K.~Jarrod Millman, St{'{e}}fan~J. van~der Walt, Ralf
  Gommers, Pauli Virtanen, David Cournapeau, Eric Wieser, Julian Taylor,
  Sebastian Berg, Nathaniel~J. Smith, Robert Kern, Matti Picus, Stephan Hoyer,
  Marten~H. van Kerkwijk, Matthew Brett, Allan Haldane, Jaime~Fern{'{a}}ndez
  del R{'{\i}}o, Mark Wiebe, Pearu Peterson, Pierre G{'{e}}rard-Marchant, Kevin
  Sheppard, Tyler Reddy, Warren Weckesser, Hameer Abbasi, Christoph Gohlke, and
  Travis~E. Oliphant.
\newblock Array programming with {NumPy}.
\newblock {\em Nature}, 585(7825):357--362, September 2020.

\bibitem{scikit-learn}
F.~Pedregosa, G.~Varoquaux, A.~Gramfort, V.~Michel, B.~Thirion, O.~Grisel,
  M.~Blondel, P.~Prettenhofer, R.~Weiss, V.~Dubourg, J.~Vanderplas, A.~Passos,
  D.~Cournapeau, M.~Brucher, M.~Perrot, and E.~Duchesnay.
\newblock Scikit-learn: Machine learning in {P}ython.
\newblock {\em Journal of Machine Learning Research}, 12:2825--2830, 2011.

\bibitem{matplot}
J.~D. Hunter.
\newblock Matplotlib: A 2d graphics environment.
\newblock {\em Computing in Science \& Engineering}, 9(3):90--95, 2007.

\bibitem{networkX}
Aric~A. Hagberg, Daniel~A. Schult, and Pieter~J. Swart.
\newblock Exploring network structure, dynamics, and function using networkx.
\newblock In Ga\"el Varoquaux, Travis Vaught, and Jarrod Millman, editors, {\em
  Proceedings of the 7th Python in Science Conference}, pages 11 -- 15,
  Pasadena, CA USA, 2008.

\bibitem{krupic_grid_2015}
Julija Krupic, Marius Bauza, Stephen Burton, Caswell Barry, and John O'Keefe.
\newblock Grid cell symmetry is shaped by environmental geometry.
\newblock {\em Nature}, 518, 2015.

\bibitem{PredictiveMap}
Kimberly~L. Stachenfeld, Matthew~M. Botvinick, and Samuel~J. Gershman.
\newblock The hippocampus as a predictive map.
\newblock {\em Nature Neuroscience}, 21(6):895--895, 2017.

\bibitem{placecells}
John O'Keefe and Neil Burgess.
\newblock Geometric determinants of the place fields hippocampal neurons.
\newblock {\em Nature}, 381:425--8, 06 1996.

\bibitem{gridmeasure}
Marianne Fyhn, Torkel Hafting, Menno~P. Witter, Edvard~I. Moser, and May-Britt
  Moser.
\newblock Grid cells in mice.
\newblock {\em Hippocampus}, 18(12):1230--1238, 2008.

\bibitem{brun2008progressive}
Vegard~Heimly Brun, Trygve Solstad, Kirsten~Brun Kjelstrup, Marianne Fyhn,
  Menno~P Witter, Edvard~I Moser, and May-Britt Moser.
\newblock Progressive increase in grid scale from dorsal to ventral medial
  entorhinal cortex.
\newblock {\em Hippocampus}, 18(12):1200--1212, 2008.

\bibitem{gerum2020sparsity}
Richard~C Gerum, Andr{\'e} Erpenbeck, Patrick Krauss, and Achim Schilling.
\newblock Sparsity through evolutionary pruning prevents neuronal networks from
  overfitting.
\newblock {\em Neural Networks}, 128:305--312, 2020.

\bibitem{sutton2018reinforcement}
Richard~S Sutton and Andrew~G Barto.
\newblock {\em Reinforcement learning: An introduction}.
\newblock MIT press, 2018.

\bibitem{goldberg1995constructions}
Adele~E Goldberg.
\newblock {\em Constructions: A construction grammar approach to argument
  structure}.
\newblock University of Chicago Press, 1995.

\bibitem{goldberg2003constructions}
Adele~E Goldberg.
\newblock Constructions: A new theoretical approach to language.
\newblock {\em Trends in cognitive sciences}, 7(5):219--224, 2003.

\bibitem{goldberg2006constructions}
Adele~E Goldberg.
\newblock {\em Constructions at work: The nature of generalization in
  language}.
\newblock Oxford University Press on Demand, 2006.

\bibitem{goldberg2019explain}
Adele Goldberg and Adele~E Goldberg.
\newblock {\em Explain me this}.
\newblock Princeton University Press, 2019.

\bibitem{diessel2019usage}
Holger Diessel, Ewa Dabrowska, and Dagmar Divjak.
\newblock Usage-based construction grammar.
\newblock {\em Cognitive linguistics}, 2:50--80, 2019.

\bibitem{diessel2020dynamic}
Holger Diessel.
\newblock A dynamic network approach to the study of syntax.
\newblock {\em Frontiers in Psychology}, page 3196, 2020.

\bibitem{levshina2021grammar}
Natalia Levshina.
\newblock The grammar network: How linguistic structure is shaped by language
  use by holger diessel.
\newblock {\em Language}, 97(4):825--830, 2021.

\bibitem{diessel2019grammar}
Holger Diessel.
\newblock {\em The grammar network}.
\newblock Cambridge University Press, 2019.

\bibitem{kuhl_brain_2010}
Patricia~K. Kuhl.
\newblock Brain {Mechanisms} in {Early} {Language} {Acquisition}.
\newblock {\em Neuron}, 67:713--727, 2010.

\bibitem{dabrowska2014implicit}
Ewa Dabrowska.
\newblock Implicit lexical knowledge.
\newblock {\em Linguistics}, 52(1):205--223, 2014.

\bibitem{bermudez2020neuroscience}
Edgar Bermudez-Contreras, Benjamin~J Clark, and Aaron Wilber.
\newblock The neuroscience of spatial navigation and the relationship to
  artificial intelligence.
\newblock {\em Frontiers in Computational Neuroscience}, 14:63, 2020.

\bibitem{kriegeskorte2018cognitive}
Nikolaus Kriegeskorte and Pamela~K Douglas.
\newblock Cognitive computational neuroscience.
\newblock {\em Nature neuroscience}, 21(9):1148--1160, 2018.

\bibitem{schilling2020intrinsic}
Achim Schilling, Richard Gerum, Alexandra Zankl, Holger Schulze, Claus Metzner,
  and Patrick Krauss.
\newblock Intrinsic noise improves speech recognition in a computational model
  of the auditory pathway.
\newblock {\em bioRxiv}, 2020.

\bibitem{krauss2018cross}
Patrick Krauss, Konstantin Tziridis, Achim Schilling, and Holger Schulze.
\newblock Cross-modal stochastic resonance as a universal principle to enhance
  sensory processing.
\newblock {\em Frontiers in neuroscience}, 12:578, 2018.

\bibitem{hassabis2017neuroscience}
Demis Hassabis, Dharshan Kumaran, Christopher Summerfield, and Matthew
  Botvinick.
\newblock Neuroscience-inspired artificial intelligence.
\newblock {\em Neuron}, 95(2):245--258, 2017.

\bibitem{marcus2018deep}
Gary Marcus.
\newblock Deep learning: A critical appraisal.
\newblock {\em arXiv preprint arXiv:1801.00631}, 2018.

\bibitem{yang2021neural}
Zijin Yang, Achim Schilling, Andreas Maier, and Patrick Krauss.
\newblock Neural networks with fixed binary random projections improve accuracy
  in classifying noisy data.
\newblock In {\em Bildverarbeitung f{\"u}r die Medizin 2021}, pages 211--216.
  Springer, 2021.

\bibitem{gerum2021integration}
Richard~C Gerum and Achim Schilling.
\newblock Integration of leaky-integrate-and-fire neurons in standard machine
  learning architectures to generate hybrid networks: A surrogate gradient
  approach.
\newblock {\em Neural Computation}, 33(10):2827--2852, 2021.

\bibitem{maier2019learning}
Andreas~K Maier, Christopher Syben, Bernhard Stimpel, Tobias W{\"u}rfl, Mathis
  Hoffmann, Frank Schebesch, Weilin Fu, Leonid Mill, Lasse Kling, and Silke
  Christiansen.
\newblock Learning with known operators reduces maximum error bounds.
\newblock {\em Nature machine intelligence}, 1(8):373--380, 2019.

\end{thebibliography}

\end{document}